\documentclass[letterpaper, 10 pt, conference]{ieeeconf}
\IEEEoverridecommandlockouts

\usepackage{cite}
\usepackage{amsmath,amssymb,amsfonts}
\usepackage{algorithmic}
\usepackage{graphicx}
\usepackage{textcomp}
\usepackage{cite}
\usepackage{multirow}
\usepackage{array}
\usepackage{hyperref}
\usepackage[font=small]{caption}
\usepackage{subcaption}
\usepackage{balance}
\usepackage[table,xcdraw]{xcolor}

\begin{document}

\makeatletter
    \newcommand{\linebreakand}{%
      \end{@IEEEauthorhalign}
      \hfill\mbox{}\par
      \mbox{}\hfill\begin{@IEEEauthorhalign}
    }
\makeatother

\def\BibTeX{{\rm B\kern-.05em{\sc i\kern-.025em b}\kern-.08em
    T\kern-.1667em\lower.7ex\hbox{E}\kern-.125emX}}

\hbadness=99999 

\title{\LARGE \bf
UAV-VLRR: Vision-Language Informed NMPC for Rapid Response in UAV Search and Rescue
}

\author{Yasheerah Yaqoot$^{*}$, Muhammad Ahsan Mustafa$^{*}$, Oleg Sautenkov, Artem Lykov, \\ Valerii Serpiva, and Dzmitry Tsetserukou
\thanks{The authors are with the Intelligent Space Robotics Laboratory, Center for Digital Engineering, Skolkovo Institute of Science and Technology. 
{\tt \{yasheerah.yaqoot, ahsan.mustafa, oleg.sautenkov, artem.lykov, valerii.serpiva, d.tsetserukou\}@skoltech.ru}}
\thanks{*These authors contributed equally to this work.}
}

\maketitle


\begin{abstract}
Emergency search and rescue (SAR) operations often require rapid and precise target identification in complex environments where traditional manual drone control is inefficient. In order to address these scenarios, a rapid SAR system, UAV-VLRR (Vision-Language-Rapid-Response), is developed in this research. This system consists of two aspects: 1) A multimodal system which harnesses the power of Visual Language Model (VLM) and the natural language processing capabilities of ChatGPT-4o (LLM) for scene interpretation. 2) A non-linear model predictive control (NMPC) with built-in obstacle avoidance for rapid response by a drone to fly according to the output of the multimodal system. This work aims at improving response times in emergency SAR operations by providing a more intuitive and natural approach to the operator to plan the SAR mission while allowing the drone to carry out that mission in a rapid and safe manner. 
When tested, our approach was faster on an average by \textbf{33.75\%} when compared with an off-the-shelf autopilot and \textbf{54.6\%} when compared with a human pilot.

Github: \href{https://github.com/ahsan-mustafa/uav-vlrr}{https://github.com/ahsan-mustafa/uav-vlrr}

Video of UAV-VLRR: \href{https://youtu.be/KJqQGKKt1xY}{https://youtu.be/KJqQGKKt1xY}
\end{abstract}
\textbf{\textit{Keywords:}} \textbf{\textit{VLM; LLM-agents; VLM-agents; UAV; Navigation; Drone; Path Planning; NMPC.}}


\section{Introduction}


Search and rescue (SAR) operations in disaster-stricken environments require fast and efficient situational assessment to locate survivors and critical infrastructure. Unmanned Aerial Vehicles (UAVs) have become vital in SAR missions due to their ability to access hard-to-reach areas, provide real-time imagery, and reduce response times \cite{uav_survey}, \cite{uav_use}. However, traditional UAV-based SAR relies heavily on manual flight control or waypoint setting. In high-stakes emergencies, the pressure can overwhelm even experienced responders, and the chaotic nature of disaster zones often leads to impaired judgment and delays in mission planning. As cognitive overload increases, critical details may be overlooked, and manual approaches can falter, as seen in our previous work FlightAR \cite{sautenkov2024flightararflightassistance}. These limitations highlight the need for an intelligent SAR system that can autonomously generate mission waypoints in complex environments with minimal human input. Furthermore, such a system must be deployed on a UAV capable of executing missions safely and rapidly.

\begin{figure}[t!]
\centering
\includegraphics[width=1\linewidth]{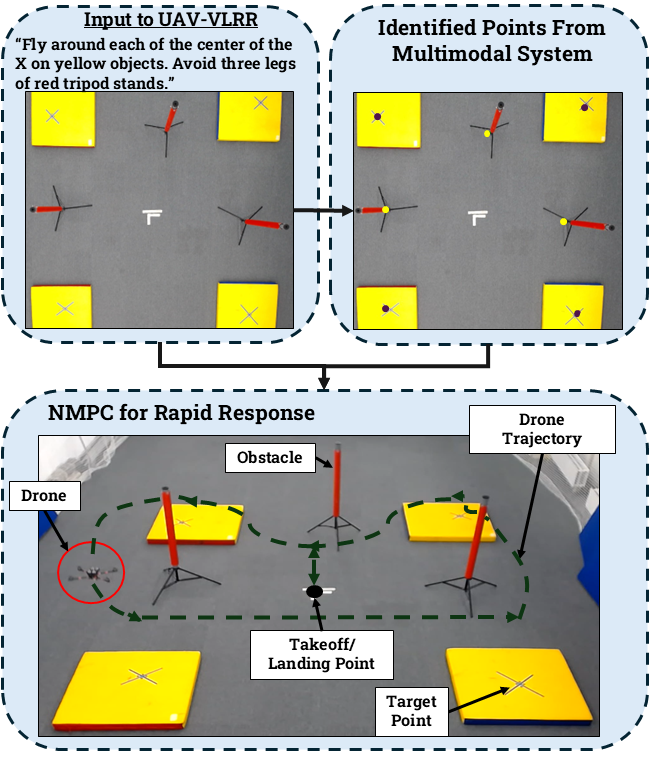} 
\caption{Illustration of the UAV-VLRR framework. The left image shows the input to the system, and the right displays the identified points by the multimodal system. Below, the NMPC guides the drone’s trajectory, ensuring obstacle avoidance and navigation to target points.}
\end{figure}

A key challenge in achieving autonomous UAV-based SAR missions lies in environmental perception and real-time decision-making. Traditional UAV mission planning techniques often depend on handcrafted obstacle maps, LiDAR-based navigation, or heuristic path-planning algorithms. While effective in structured environments, these approaches struggle to adapt to the unpredictable nature of disaster zones where obstacles, such as collapsed buildings, debris, and vegetation, are constantly changing. To address these limitations, there is a need for a system that can autonomously interpret aerial imagery, extract relevant information, and generate actionable flight paths in real time. One of our previous research \cite{sautenkov2025uavvlavisionlanguageactionlargescale} involves a UAV-VLA framework built on this concept.

In this work, we build on the UAV-VLA framework \cite{sautenkov2025uavvlavisionlanguageactionlargescale} by integrating its capability of interpreting aerial images with the agile control of a quadrotor resulting in quick coverage of the destination points given by the multimodal system. Our contributions are as follows:

\begin{itemize}
    \item We introduce the \textbf{UAV-VLRR} framework, combining the multimodal Vision-Language interpretation of aerial images with rapid control. 
    \item We apply a point-to-point Non-linear Model Predictive Control (NMPC) control scheme with built in obstacle avoidance to ensure safe and rapid UAV response in complex environments.
    \item We demonstrate that our framework outperforms other traditional approaches in the field of drone search and rescue.
\end{itemize}

\section{Related Work}

\subsection{Multimodal Vision-Language Approaches for Robotic Systems}
The introduction of Vision Transformers (ViTs) \cite{visiontransformerdosovitskiy2021imageworth16x16words}, \cite{CLIPradford2021learningtransferablevisualmodels} marked a pivotal shift in the development of models capable of integrating various input and output modalities, including text, images, and video. This progress laid the foundation for models such as OpenAI's ChatGPT-4 Omni \cite{openai2024gpt4technicalreport}, which can perform real-time reasoning across multiple modalities, enhancing multimodal interactions. In the robotics domain, the Allen Institute for AI introduced the Molmo model, which uses image-text pairs to locate objects in response to user requests \cite{deitke2024molmopixmoopenweights}, further advancing the integration of vision and language in robotic systems.

Vision-Language models have also been applied in UAV control. Sautenkov et al. \cite{sautenkov2024flightararflightassistance} improved drone surveillance using multiple video streams and object detection, although manual operation was still required. The Google DeepMind’s RT-2 \cite{brohan2023rt2visionlanguageactionmodelstransfer} introduced models advanced this field by enabling direct robot control from multimodal sensory inputs. The UAV-VLA framework, as presented in \cite{sautenkov2025uavvlavisionlanguageactionlargescale}, takes these advancements further by using multimodal systems to generate actionable mission paths through text-image pairs. This approach underscores the critical role of vision and language integration in a variety of robotic applications, particularly in tasks that require real-time environmental understanding.

Further expanding this line of research, UAV-CodeAgents \cite{sautenkov2025uavcodeagentsscalableuavmission} introduced reasoning step for navigation, and UAV-VLPA* \cite{sautenkov2025uavvlpavisionlanguagepathactionoptimalroute} introduced global route optimization by combining TSP and A* path planning, significantly reducing trajectory lengths in large-scale UAV missions. RaceVLA \cite{serpiva2025racevlavlabasedracingdrone} and CognitiveDrone \cite{lykov2025cognitivedronevlamodelevaluation} applied VLA models to racing drones and drone reasoning, producing real-time velocity and yaw commands from FPV video and language inputs, and achieving human-like decision-making in dynamic racing environments. These advances highlight the growing versatility of vision-language-action systems in aerial robotics.

Building on these developments, UAV-VLRR focuses on real-time mission execution in cluttered environments by integrating semantic understanding with onboard NMPC, enabling fast and safe UAV operation for critical applications such as search and rescue.

\begin{figure*}[t!]
\centering
\includegraphics[width=1\linewidth]{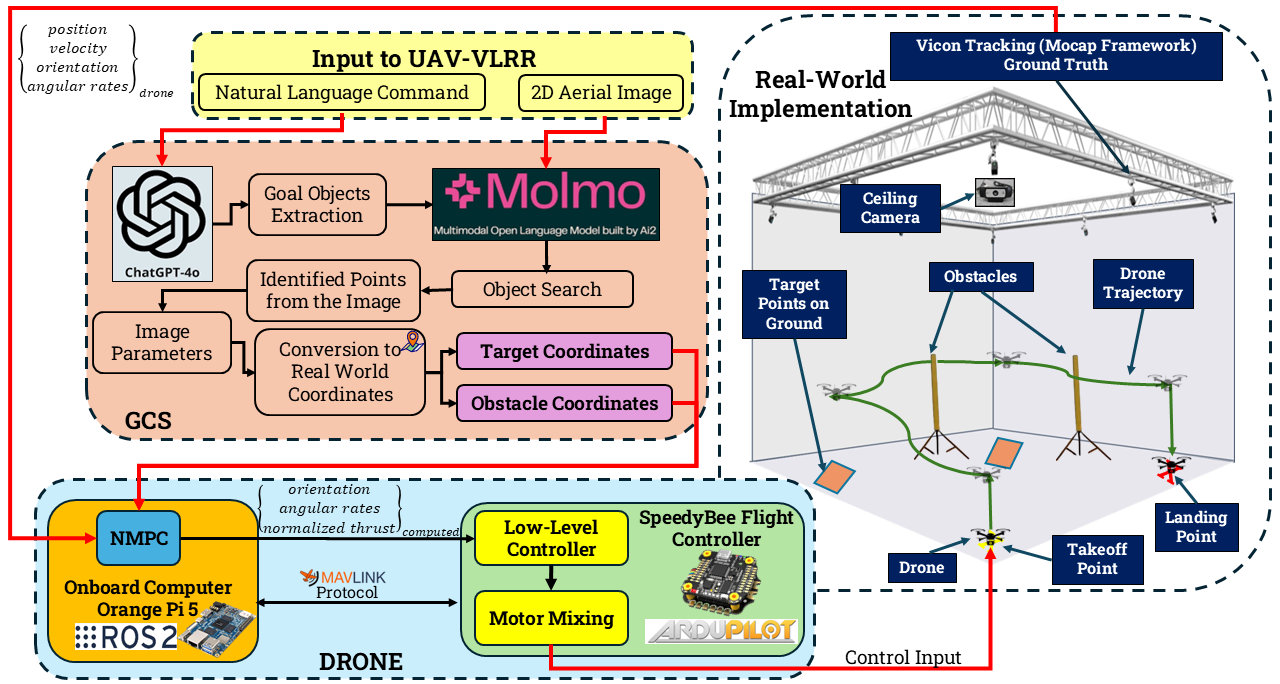} 
\caption{System architecture of the UAV-VLRR framework.}
\label{fig:architecture}
\end{figure*}

\subsection{Safe Agile Control for Drones}
The importance of NMPC for agile drones can be seen in its use by the drone racing team at ETH Zurich, which are the best in the world for high-speed drone control. They have used NMPC in many of their works \cite{differential_nmpc}, \cite{mpcc}, \cite{adaptive_mpc}, \cite{data_driven_mpc}. Sun et al. \cite{differential_nmpc} did a comparative study between NMPC and DFBC in which NMPC outperformed DFBC in terms of tracking dynamically infeasible trajectories, although it required significantly higher computational resources, which could be a bottleneck in real-time systems. This study provided critical insights into the trade-offs between computational efficiency and control performance. Romero et al. \cite{mpcc} tackled the agile drone problem by using a model predictive contouring control approach that resulted in time-optimal trajectories in real-time with effective high-speed control. However, their work was computationally expensive and they have stated that the controller was not run onboard the drone but rather on an external computer. Hanover et al. \cite{adaptive_mpc} used an adaptive MPC approach by cascading the MPC with an \(L\)1-Adaptive Controller. This resulted in immediate model mismatches and disturbances very effectively, but they have stated that there is potential for violating actuator constraints due to the inner cascaded loop. Torrente et al. \cite{data_driven_mpc} used a data-driven mpc approach by modeling aerodynamic forces using Gaussian processes, but the controller was run off-board here as well. In \cite{mpc_obs}, Ramezani et al. implemented obstacle avoidance in their MPC framework along with using long-short-term memory for states predicition. However, this work was performed only in simulation while using a simplified 3-DOF kinematics drone model. Moreover, one of our previous works, SafeSwarm \cite{safeswarm}, worked on safe drone landings in crowded areas, which is an important factor in crowded emergency scenarios. 

In order for the drone to fly at high speeds in a satisfactory manner with a minimal dynamics model (without any compensation for drag or aerodynamic mismatches), a point-to-point NMPC technique is utilized in this paper, unlike most of the work mentioned in the literature, which focus on first an external trajectory generation module and then a trajectory following module. This approach enables the drone to fly properly despite the model mismatches since now it does not have strict constraints of tracking a given trajectory. Moreover, this also simplifies the computational need and hence is able to be deployed on an onboard computer like an OrangePi in our case. This approach is also advantageous in the sense that it only requires the target points and the obstacle points, which perfectly fits in the pipeline when cascading with the multimodal system. 

\section{System Overview}

\subsection{Vision-Language Integration for Accurate Object Identification}

In this work, a multimodal system comprising a Large Language Model (LLM) and a Vision-Language Model (VLM) is employed to enhance environmental understanding, as illustrated in Fig.~\ref{fig:architecture}. ChatGPT-4o serves as the LLM agent, responsible for extracting goal objects, specifically “target points” and “obstacles”. The quantized Molmo-7B-D BnB 4-bit model \cite{molmo_quantized} is utilized as the VLM agent for image processing and goal object identification.

The system processes an image-text pair as input, which is handled by both the LLM and VLM agents. The image-text pair processing can be mathematically represented as:

\begin{equation}
    \mathcal{C} = f_{\text{LLM}, \text{VLM}}\left(I, T\right),
\end{equation}
where \(I\) is the input image and \(T\) is the input text, and \(\mathcal{C}\) represents the output coordinates of the identified goal objects.

Once the goal objects are identified, their pixel coordinates are mapped onto the image and converted into real-world coordinates using image metadata. Specifically, the real-world coordinates are computed based on the camera's height and field-of-view (FoV) parameters. The horizontal and vertical real-world dimensions in meters are first calculated from the diagonal FoV and camera height using the following formulas:

\begin{equation}
    \text{Real width (m)} = 2 \cdot h_{\text{camera}} \cdot \tan\left(\frac{\theta_{\text{horizontal}}}{2}\right),
\end{equation}

\begin{equation}
    \text{Real height (m)} = 2 \cdot h_{\text{camera}} \cdot \tan\left(\frac{\theta_{\text{vertical}}}{2}\right),
\end{equation}
where \( h_{\text{camera}} \) is the camera's height above the ground, and \( \theta_{\text{horizontal}} \) and \( \theta_{\text{vertical}} \) are the horizontal and vertical FoVs, respectively, which are derived from the diagonal FoV and aspect ratio of the camera.

Once the real-world dimensions of the image are known, they are used to define the Cartesian coordinate bounds for the image. The normalized pixel coordinates of detected objects are mapped into real-world Cartesian coordinates by scaling them according to the image's real-world dimensions. These real-world coordinates represent the target points and obstacles, which are then passed to the NMPC for task execution.

\subsection{NMPC for Rapid Response}
The non-linear model predictive control in this research follows a point-to-point architecture. In addition, the objective function has a penalty term associated with the obstacle points received from the multimodal system. This NMPC setup enables the controller to not depend on any external trajectory generation technique. The NMPC finds the optimal trajectory and the set of control inputs designed for rapid control. 

The dynamics of the quadrotor system is governed by 13 states where
\( p_W = \left[ p_x, p_y, p_z \right]^T \) are the position coordinates in the world frame,
\( v_W = \left[ v_x, v_y, v_z \right]^T \) are the linear velocity components in the world frame, \( q_B = \left( q_{\omega}, q_x, q_y, q_z \right)^T \) are the quaternions for the orientation of the drone's body, and finally \( \omega_B = \left[ \omega_x, \omega_y, \omega_z \right]^T \) are the body angular rates. 

\[
\dot{x} = \left[ \begin{array}{c}
\dot{p}_W \\
\dot{v}_W \\
\dot{q}_B \\
\dot{\omega}_B
\end{array} \right]
= \left[ \begin{array}{c}
{v_W} \\
R(q) \frac{T_B}{m} + g\\
\frac{1}{2} q_{\omega B} \cdot q_B \\
J^{-1} \left( \tau_B - \omega_B \times J \omega_B \right)
\end{array} \right],
\]
where \(R(q)\) is the quaternion rotational matrix, \( T_B = \left[ 0, 0, \sum_{i=1}^4 T_i \right]^T \) is the total thrust, \(g\) is the gravitational vector \( g = \left[ 0, 0, 9.81 \right]^T \), \( J = \text{diag}(J_x, J_y, J_z) \) is the diagonal of the inertia matrix and \( q_{\omega} = \left( 0, \omega_x, \omega_y, \omega_z \right)^T \) is the angular velocity quaternion.

The drone's body torque matrix is according to the free body diagram shown in Fig.~\ref{fig:drone_cs}. The body torque matrix comes out to be:

\[
\tau_B = \begin{bmatrix}
-l_y & l_y & l_y & -l_y \\
-l_x & -l_x & l_x & l_x \\
k_t & -k_t & k_t & -k_t
\end{bmatrix}
\begin{bmatrix}
u_1 \\
u_2 \\
u_3 \\
u_4
\end{bmatrix} ,
\]
where \( u_1, u_2, u_3, u_4\) are the input motor forces, \( l_x, l_y\) are the distances to the $x$-axis and $y$-axis, respectively and, \( k_t\) is the torque constant.

\begin{figure}[htbp]
\centering
\includegraphics[width=0.7\linewidth]{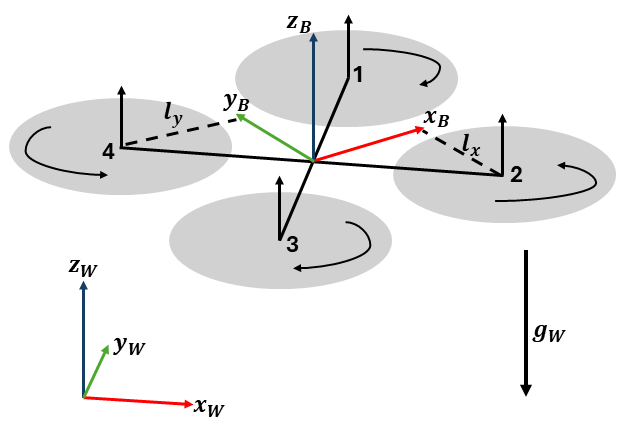} 
\caption{Drone free-body diagram.}
\label{fig:drone_cs}
\end{figure}

In order to form a discretized nonlinear optimal control problem, the Runge-Kutta method of 4th order was used:

\[
x(k+1) = f_{\text{RK4}}(x(k), u(k), \delta t)
\]

The NMPC was formulated in a multiple shooting scheme. The constructed optimization problem is as below:

\begin{equation}
l(x,u) = \lVert x_u - x_r \rVert_Q^2 + \lVert u \rVert_R^2 + Penalty_{Obs},
\label{eq:oldL}
\end{equation}
\begin{equation}
\min_{u} J(x,u) = \sum_{k=0}^{N-1} l(x_u(k), u(k)),
\end{equation}

subject to:
\[ x(k+1) = f_{\text{RK4}}(x(k), u(k), \delta t), \]
\[ x_u(0) = x_0, \]
\[ u_{min} \leq u(k) \leq u_{max}, \quad \forall k \in [0, N-1], \]
\[ x(k) \in X, \quad \forall k \in [0, N] \]
\[ Obstacle_{x,y}\]

The system was discretized into a prediction horizon of \( N\) steps with a step horizon of \( T\) between each step. The control problem is iteratively solved in real time onboard the drone using CasADi \cite{Andersson2019}.

\section{Experimental Setup}

The UAV-VLRR framework was tested inside the drone arena of the Intelligent Space Robotics Lab at Skoltech. The command given to the system was: \textit{“Fly around each of the center of the X on yellow objects. Avoid three legs of red tripod stands.”} The system was tested under various conditions, with the following three experiments performed:

\begin{itemize}
    \item \textbf{Exp 1:} The drone flies to the target points using the UAV-VLRR framework.  
    \item \textbf{Exp 2:} The drone flies to the target points using an off-the-shelf autopilot.
    \item \textbf{Exp 3:} A human drone pilot is shown the picture and then flies around the target points while having access to a belly-mounted camera on the drone.
\end{itemize}

There were two different scenarios in which all the three experiments were conducted:

\begin{itemize}
    \item \textbf{Scene 1:} There were three target points (X marked on yellow objects) and two obstacles (red tripod stands).
    \item \textbf{Scene 2:} There were four target points (X marked on yellow objects) and three obstacles (red tripod stands).
\end{itemize}

The multimodal system ran on a remote server with an RTX 4090 GPU (24GB VRAM) and an Intel Core i9-13900K processor. During the experiment, the drone sent the image and command to the server, which returned the target and obstacle coordinates. The same 2D aerial image and prompt were also given to the human pilot for comparison. Images used in Scene 1 and Scene 2 are shown in Fig.~\ref{fig:scenes}.

\begin{figure}[h]
\centering
\begin{subfigure}{.24\textwidth}
  \centering
  \includegraphics[width=.8\linewidth]{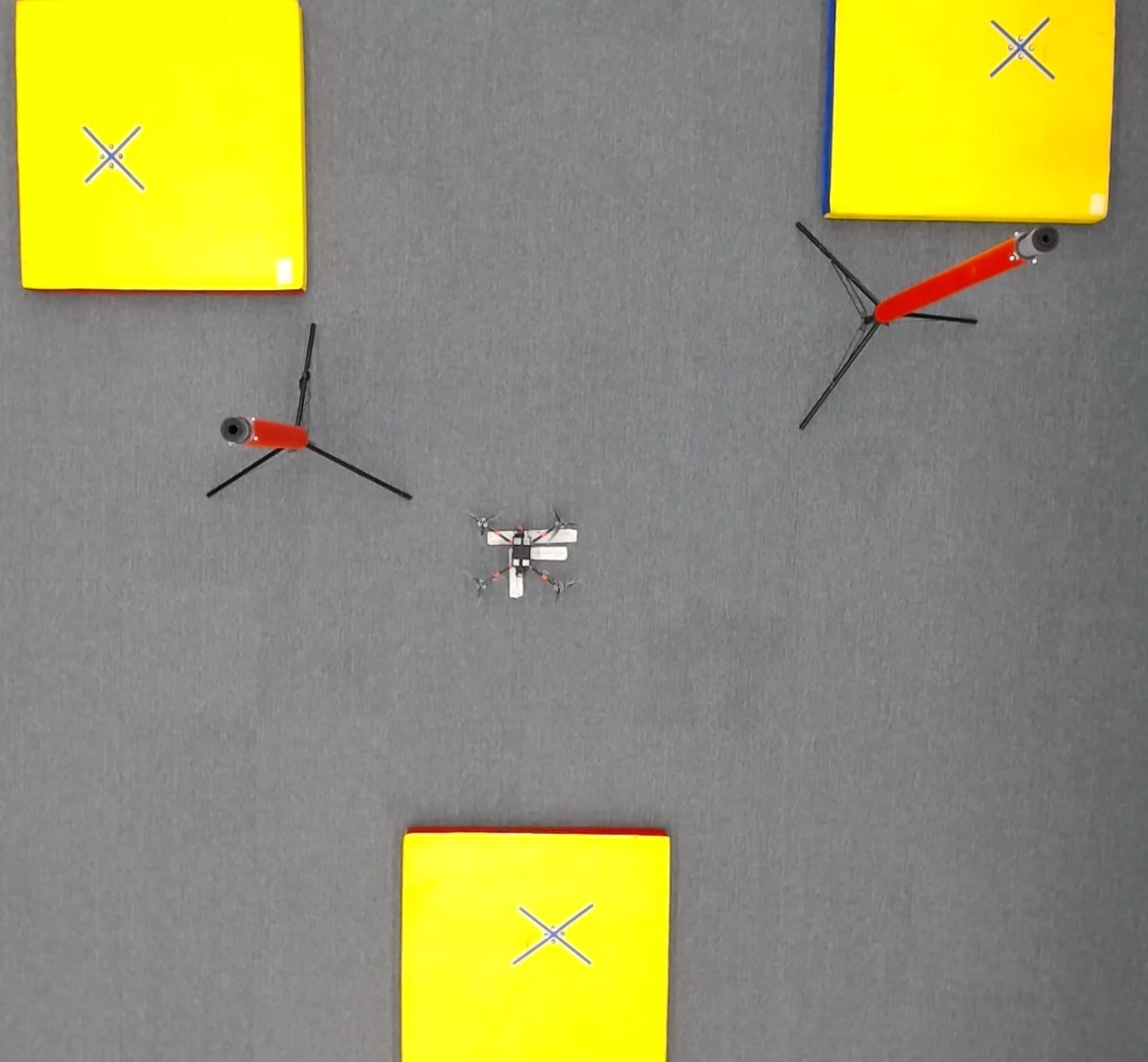}
  \caption{Scene 1.}
  \label{fig:sub1}
\end{subfigure}%
\begin{subfigure}{.24\textwidth}
  \centering
  \includegraphics[width=.88\linewidth]{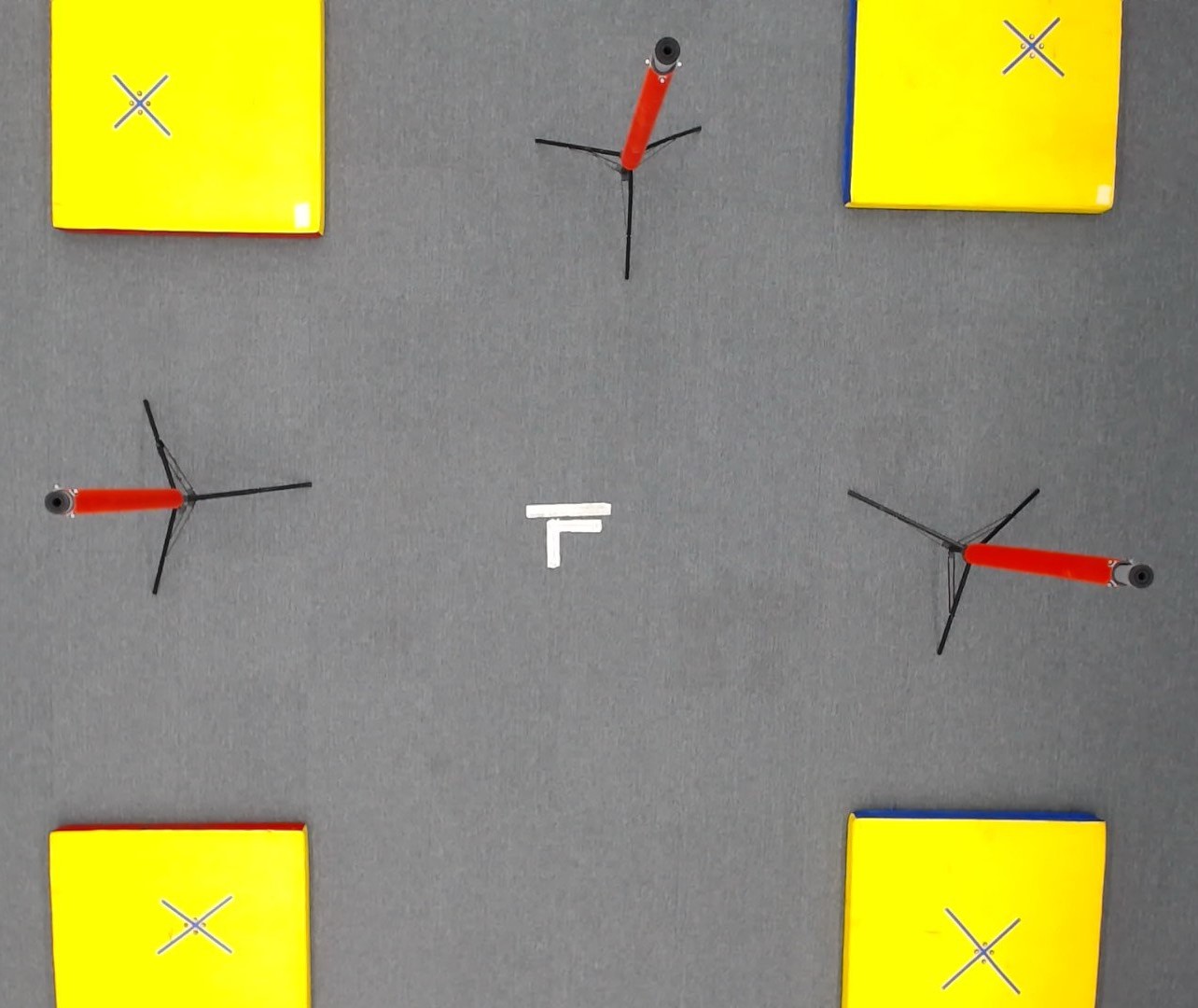}
  \caption{Scene 2.}
  \label{fig:sub2}
\end{subfigure}
\caption{Scenes used in the experiment with target points (X on yellow objects) and obstacles (red tripod stands).}
\label{fig:scenes}
\end{figure}

Each experiment was timed to assess the speed of execution. For the UAV-VLRR framework, the timing started as soon as the code was launched, while for the human pilot, the timing started as soon as the image was shown. The goal of all experiments was to have the drone fly to the required points and avoid obstacles in the shortest time possible.

\section{Experimental Results}

\subsection{Multimodal System Results}
The results obtained from the multimodal system were compared to the Vicon data for the target points and obstacles in both scenes. The identified images for Scene 1 and Scene 2 can be seen in Fig.~\ref{fig:vlm_results}, which illustrates the target points and obstacles detected by the system.

\begin{figure}[h]
\centering
\begin{subfigure}{.24\textwidth}
  \centering
  \includegraphics[width=.86\linewidth]{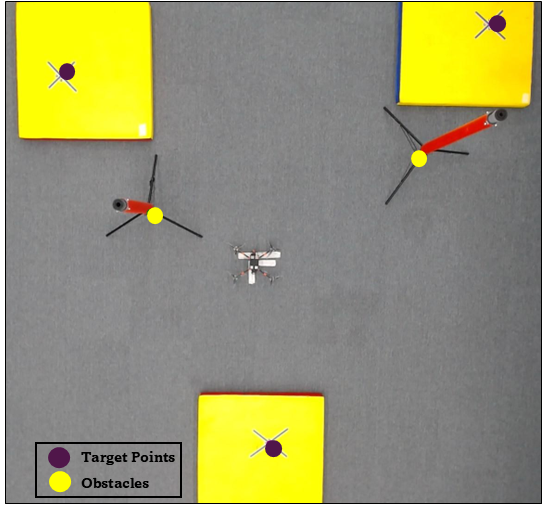}
  \caption{Results for Scene 1.}
  \label{fig:sub1}
\end{subfigure}%
\begin{subfigure}{.24\textwidth}
  \centering
  \includegraphics[width=.92\linewidth]{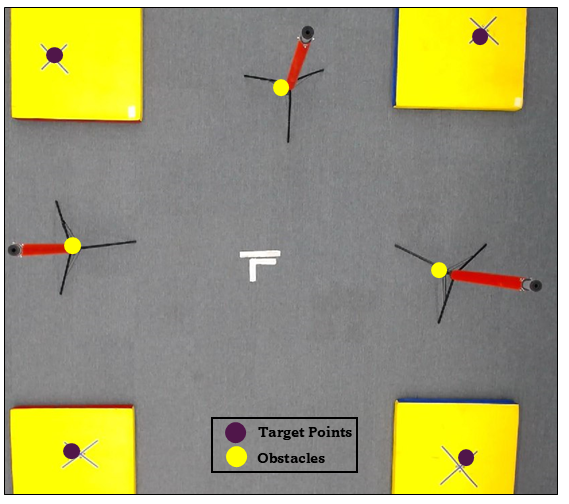}
  \caption{Results for Scene 2.}
  \label{fig:sub2}
\end{subfigure}
\caption{Identified target points and obstacles from the multimodal system for the given image-text pairs in both scenes.}
\label{fig:vlm_results}
\end{figure}

Tables~\ref{table:s1_vlm} and \ref{table:s2_vlm} present the ground truth values alongside the multimodal system’s detected points, as well as the corresponding accuracy for Scene 1 and Scene 2, respectively. For this analysis, an identification was considered accurate if the detected point was within a 25 cm radius of the actual object. This threshold accounts for the safety radius and obstacle gain applied in the NMPC to ensure safety during navigation.

It is worth noting that the image was not captured from a very high altitude, which may have resulted in some distortion or skewing. As a result, there was a higher error in the identification of some of the objects, but this is expected due to the imaging conditions at the time of capture.

\begin{table}[h]
\caption{\textsc{Comparison of Vicon (Ground Truth) and Multimodal System for Scene 1 with Error Values}}
\centering
\begin{tabular}{|ccccc|}
\hline
\rowcolor[HTML]{FFE972} 
\multicolumn{5}{|c|}{\textbf{Scene 1}}                                                                                                                                                                                                                                                                                \\ \hline
\multicolumn{2}{|c|}{}                                                     & \multicolumn{1}{c|}{\textbf{\begin{tabular}[c]{@{}c@{}}Vicon\\ coordinates\end{tabular}}} & \multicolumn{1}{c|}{\textbf{\begin{tabular}[c]{@{}c@{}}Multimodal\\ coordinates\end{tabular}}} & \textbf{\begin{tabular}[c]{@{}c@{}}Error\\ (cm)\end{tabular}} \\ \hline
\multicolumn{1}{|c|}{\multirow{2}{*}{Target 1}}   & \multicolumn{1}{c|}{X} & \multicolumn{1}{c|}{-1.42}                                                       & \multicolumn{1}{c|}{-1.27}                                                            & 15                                                   \\ \cline{2-5} 
\multicolumn{1}{|c|}{}                            & \multicolumn{1}{c|}{Y} & \multicolumn{1}{c|}{-1.39}                                                       & \multicolumn{1}{c|}{-1.25}                                                            & 14                                                   \\ \hline
\multicolumn{1}{|c|}{\multirow{2}{*}{Target 2}}   & \multicolumn{1}{c|}{X} & \multicolumn{1}{c|}{1.43}                                                        & \multicolumn{1}{c|}{1.34}                                                             & 9                                                    \\ \cline{2-5} 
\multicolumn{1}{|c|}{}                            & \multicolumn{1}{c|}{Y} & \multicolumn{1}{c|}{0.13}                                                        & \multicolumn{1}{c|}{0.18}                                                             & 5                                                    \\ \hline
\multicolumn{1}{|c|}{\multirow{2}{*}{Target 3}}   & \multicolumn{1}{c|}{X} & \multicolumn{1}{c|}{-1.76}                                                       & \multicolumn{1}{c|}{-1.60}                                                            & 16                                                   \\ \cline{2-5} 
\multicolumn{1}{|c|}{}                            & \multicolumn{1}{c|}{Y} & \multicolumn{1}{c|}{1.82}                                                        & \multicolumn{1}{c|}{1.73}                                                             & 9                                                    \\ \hline
\multicolumn{1}{|c|}{\multirow{2}{*}{Obstacle 1}} & \multicolumn{1}{c|}{X} & \multicolumn{1}{c|}{-0.41}                                                       & \multicolumn{1}{c|}{-0.28}                                                            & 13                                                   \\ \cline{2-5} 
\multicolumn{1}{|c|}{}                            & \multicolumn{1}{c|}{Y} & \multicolumn{1}{c|}{-0.72}                                                       & \multicolumn{1}{c|}{-0.65}                                                            & 7                                                    \\ \hline
\multicolumn{1}{|c|}{\multirow{2}{*}{Obstacle 2}} & \multicolumn{1}{c|}{X} & \multicolumn{1}{c|}{-0.85}                                                       & \multicolumn{1}{c|}{-0.66}                                                            & 19                                                   \\ \cline{2-5} 
\multicolumn{1}{|c|}{}                            & \multicolumn{1}{c|}{Y} & \multicolumn{1}{c|}{1.31}                                                        & \multicolumn{1}{c|}{1.20}                                                             & 11                                                   \\ \hline
\end{tabular}
\label{table:s1_vlm}
\end{table}

\begin{table}[]
\caption{\textsc{Comparison of Vicon (Ground Truth) and Multimodal System for Scene 2 with Error Values} }
\centering
\begin{tabular}{|ccccc|}
\hline
\rowcolor[HTML]{FFE972} 
\multicolumn{5}{|c|}{\textbf{Scene 2}}                                                                                                                                                                                                                                                                                \\ \hline
\multicolumn{2}{|c|}{}                                                     & \multicolumn{1}{c|}{\textbf{\begin{tabular}[c]{@{}c@{}}Vicon\\ coordinates\end{tabular}}} & \multicolumn{1}{c|}{\textbf{\begin{tabular}[c]{@{}c@{}}Multimodal\\ coordinates\end{tabular}}} & \textbf{\begin{tabular}[c]{@{}c@{}}Error\\ (cm)\end{tabular}} \\ \hline
\multicolumn{1}{|c|}{\multirow{2}{*}{Target 1}}   & \multicolumn{1}{c|}{X} & \multicolumn{1}{c|}{-1.42}                                                       & \multicolumn{1}{c|}{-1.42}                                                            & 0                                                    \\ \cline{2-5} 
\multicolumn{1}{|c|}{}                            & \multicolumn{1}{c|}{Y} & \multicolumn{1}{c|}{-1.39}                                                       & \multicolumn{1}{c|}{-1.38}                                                            & 1                                                    \\ \hline
\multicolumn{1}{|c|}{\multirow{2}{*}{Target 2}}   & \multicolumn{1}{c|}{X} & \multicolumn{1}{c|}{1.56}                                                        & \multicolumn{1}{c|}{1.44}                                                             & 12                                                   \\ \cline{2-5} 
\multicolumn{1}{|c|}{}                            & \multicolumn{1}{c|}{Y} & \multicolumn{1}{c|}{-1.43}                                                       & \multicolumn{1}{c|}{-1.25}                                                            & 18                                                   \\ \hline
\multicolumn{1}{|c|}{\multirow{2}{*}{Target 3}}   & \multicolumn{1}{c|}{X} & \multicolumn{1}{c|}{1.73}                                                        & \multicolumn{1}{c|}{1.50}                                                             & 23                                                   \\ \cline{2-5} 
\multicolumn{1}{|c|}{}                            & \multicolumn{1}{c|}{Y} & \multicolumn{1}{c|}{1.70}                                                        & \multicolumn{1}{c|}{1.60}                                                             & 10                                                   \\ \hline
\multicolumn{1}{|c|}{\multirow{2}{*}{Target 4}}   & \multicolumn{1}{c|}{X} & \multicolumn{1}{c|}{-1.76}                                                       & \multicolumn{1}{c|}{-1.55}                                                            & 21                                                   \\ \cline{2-5} 
\multicolumn{1}{|c|}{}                            & \multicolumn{1}{c|}{Y} & \multicolumn{1}{c|}{1.82}                                                        & \multicolumn{1}{c|}{1.70}                                                             & 12                                                   \\ \hline
\multicolumn{1}{|c|}{\multirow{2}{*}{Obstacle 1}} & \multicolumn{1}{c|}{X} & \multicolumn{1}{c|}{-1.29}                                                       & \multicolumn{1}{c|}{-1.16}                                                            & 13                                                   \\ \cline{2-5} 
\multicolumn{1}{|c|}{}                            & \multicolumn{1}{c|}{Y} & \multicolumn{1}{c|}{0.28}                                                        & \multicolumn{1}{c|}{0.26}                                                             & 2                                                    \\ \hline
\multicolumn{1}{|c|}{\multirow{2}{*}{Obstacle 2}} & \multicolumn{1}{c|}{X} & \multicolumn{1}{c|}{0.08}                                                        & \multicolumn{1}{c|}{0.14}                                                             & 6                                                    \\ \cline{2-5} 
\multicolumn{1}{|c|}{}                            & \multicolumn{1}{c|}{Y} & \multicolumn{1}{c|}{1.57}                                                        & \multicolumn{1}{c|}{1.38}                                                             & 19                                                   \\ \hline
\multicolumn{1}{|c|}{\multirow{2}{*}{Obstacle 3}} & \multicolumn{1}{c|}{X} & \multicolumn{1}{c|}{-0.15}                                                       & \multicolumn{1}{c|}{-0.03}                                                            & 12                                                   \\ \cline{2-5} 
\multicolumn{1}{|c|}{}                            & \multicolumn{1}{c|}{Y} & \multicolumn{1}{c|}{-1.35}                                                       & \multicolumn{1}{c|}{-1.25}                                                            & 10                                                   \\ \hline
\end{tabular}
\label{table:s2_vlm}
\end{table}

\subsection{Mission Results}

The times for each of the experiments are listed in Table~\ref{table:flight_times}. For scene 1, it can be observed that experiment 1 achieved the fastest time to complete the mission in scene 1 with 28 seconds, while experiment 3 took the most time to complete the mission with 57 seconds. Experiment 1 was 30\% faster than experiment 2 and 50.9\% faster than experiment 3.  In scene 2, once again experiment 1 achieved the fastest time for mission completion with 30 seconds while experiment 3 was again the slowest with 72 seconds. in this scene, Experiment 1 was 37.5\% faster than experiment 2 and 58.3\% faster than experiment 3. It can be deduced from the experiments that the multimodal setup (Experiments 1 and 2) outperformed the human pilots. Moreover, during manual flights, the human pilot was more prone to crashing into the obstacles. When comparing the flight times of experiment 1 and 2, it was evident that the custom NMPC was able to complete the mission faster than the off-the-shelf autopilot and was consistently better.

\begin{table}[htbp]
\caption{\textsc{Flight Results}}
\begin{center}
\begin{tabular}{|c|c|c|c|}
\hline
\rowcolor[HTML]{FFE972} 
\textbf{ } & \textbf{Exp 1} & \textbf{Exp 2} & \textbf{Exp 3} \\
\hline
Time Taken for Scene 1 (s)   & 28   &  40   &  57 \\
\hline
Time Taken for Scene 2 (s) & 30   & 48   &  72 \\
\hline
\end{tabular}
\label{table:flight_times}
\end{center}
\end{table}

The flight trajectories for scene 1 and 2 experiments are shown in Fig.~\ref{fig:scene1_results} and Fig.~\ref{fig:scene2_results} respectively.

\begin{figure}[h!]
\centering
\begin{subfigure}{.24\textwidth}
  \centering
  \includegraphics[width=1\linewidth]{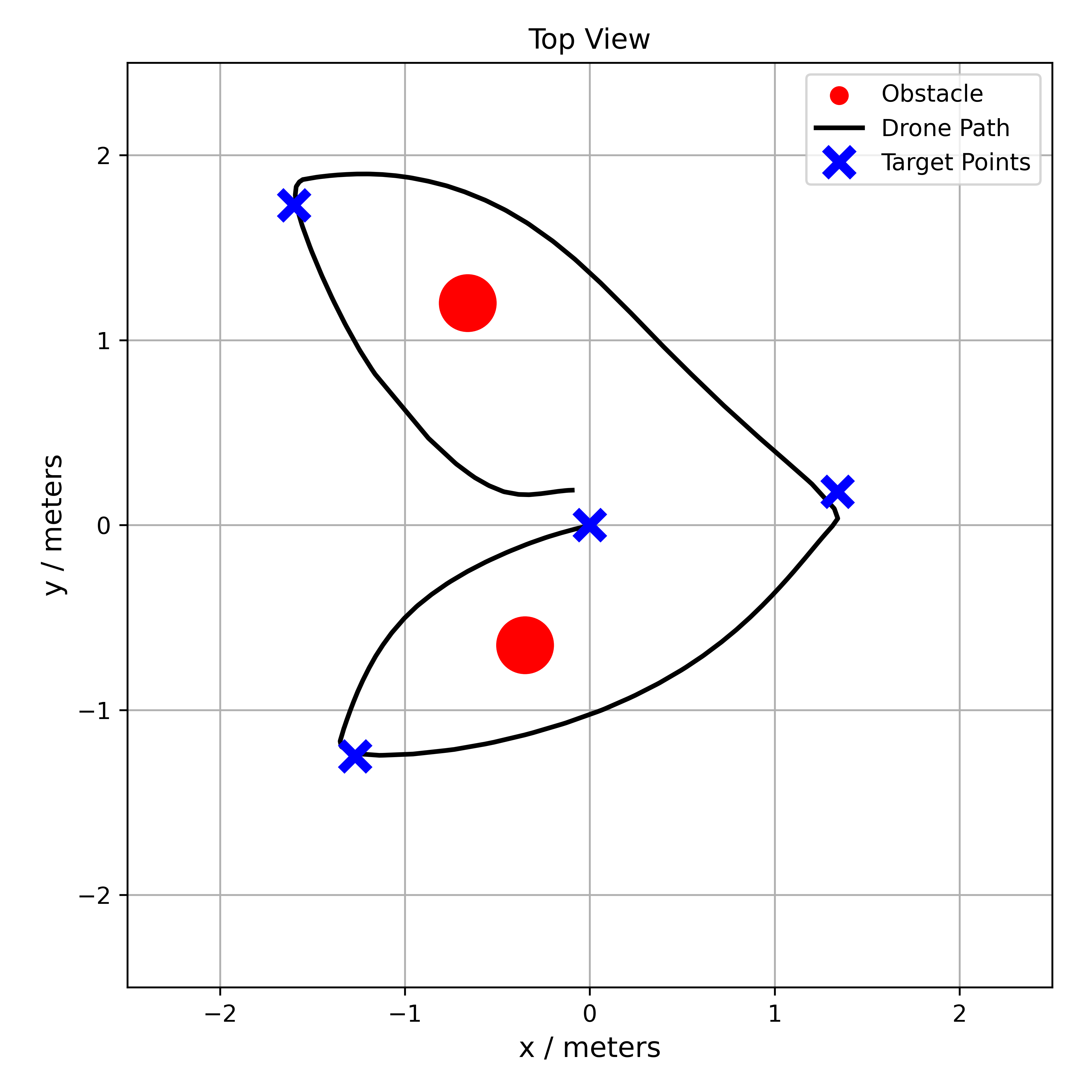}
  \caption{Scene 1 - UAV-VLRR.}
  \label{fig:sub1}
\end{subfigure}%
\begin{subfigure}{.24\textwidth}
  \centering
  \includegraphics[width=1\linewidth]{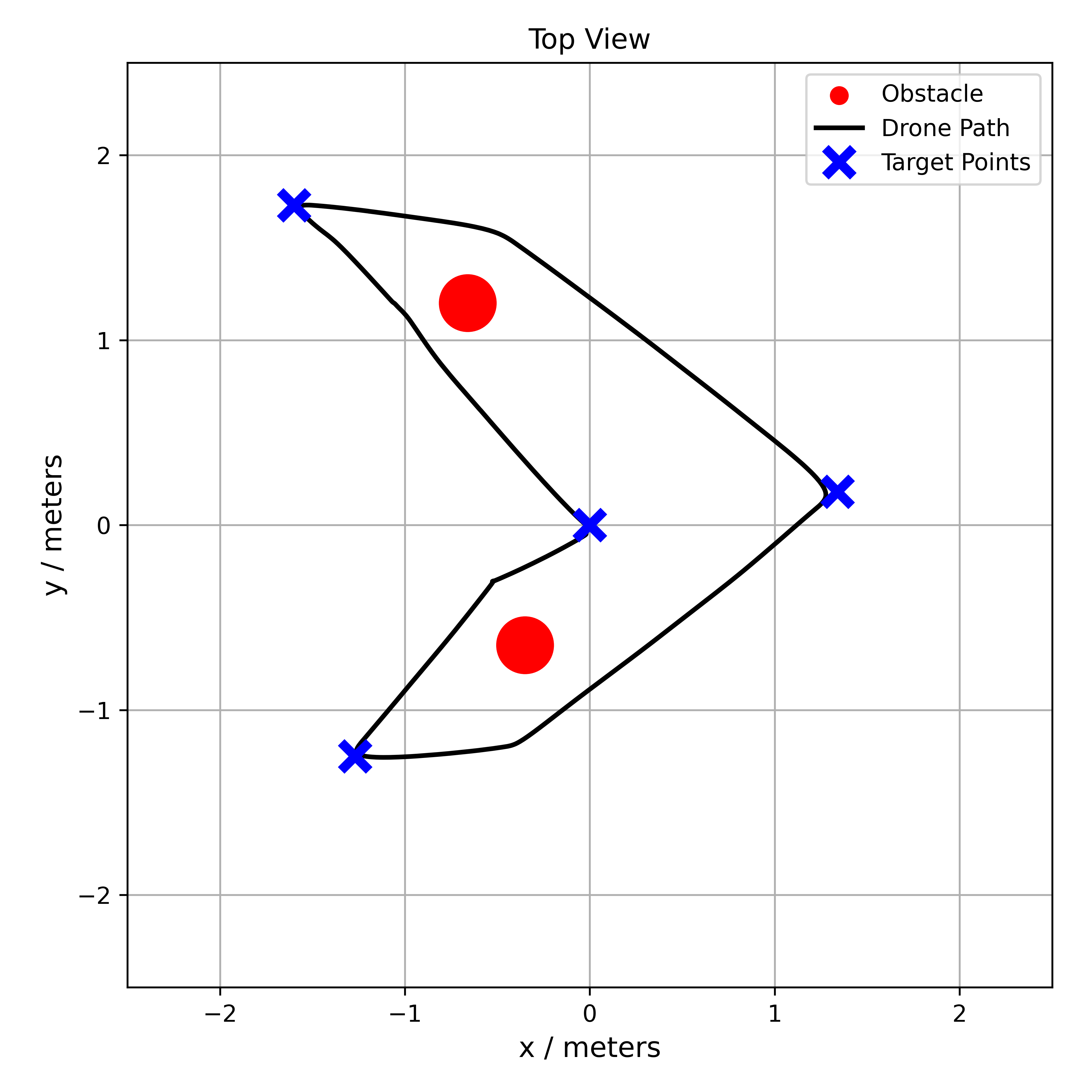}
  \caption{Scene 1 - Off-the-Shelf Autopilot.}
  \label{fig:sub2}
\end{subfigure}
\begin{subfigure}{.24\textwidth}
  \centering
  \includegraphics[width=1\linewidth]{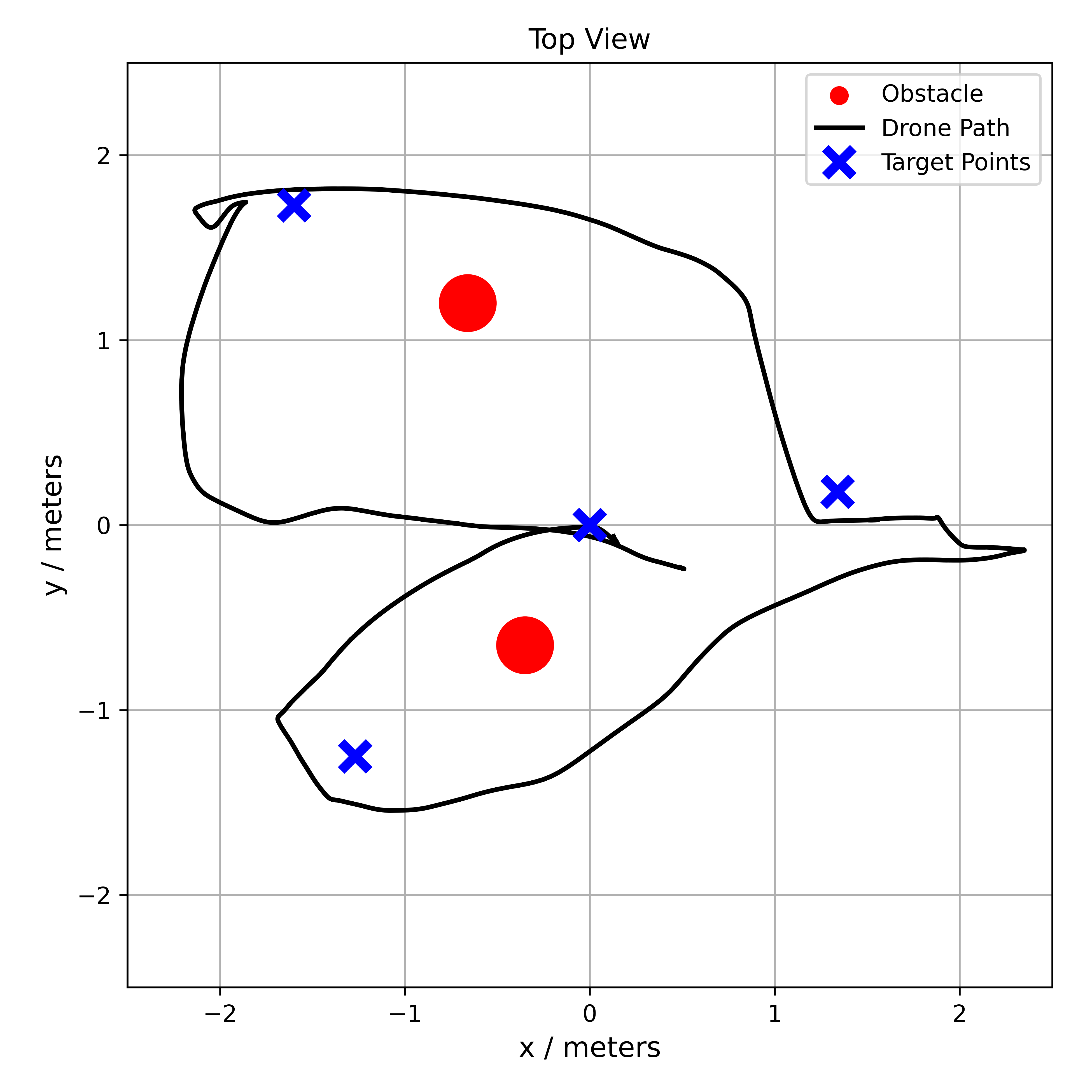}
  \caption{Scene 1 - Human Pilot.}
  \label{fig:sub2}
\end{subfigure}
\caption{Flight paths for the 3 experiments for Scene 1.}
\label{fig:scene1_results}
\end{figure}

\begin{figure}[h!]
\centering
\begin{subfigure}{.24\textwidth}
  \centering
  \includegraphics[width=1\linewidth]{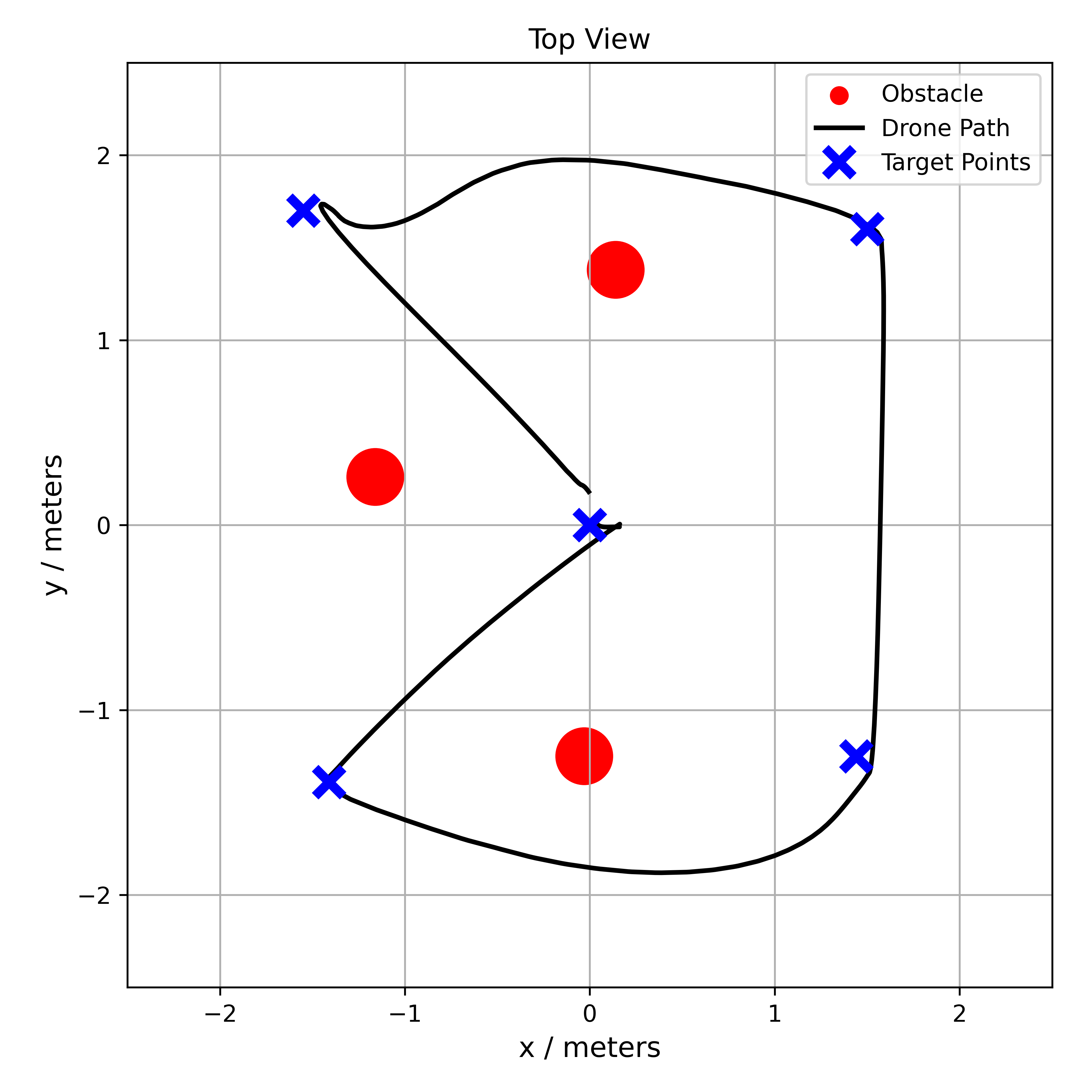}
  \caption{Scene 2 - UAV-VLRR.}
  \label{fig:sub1}
\end{subfigure}%
\begin{subfigure}{.24\textwidth}
  \centering
  \includegraphics[width=1\linewidth]{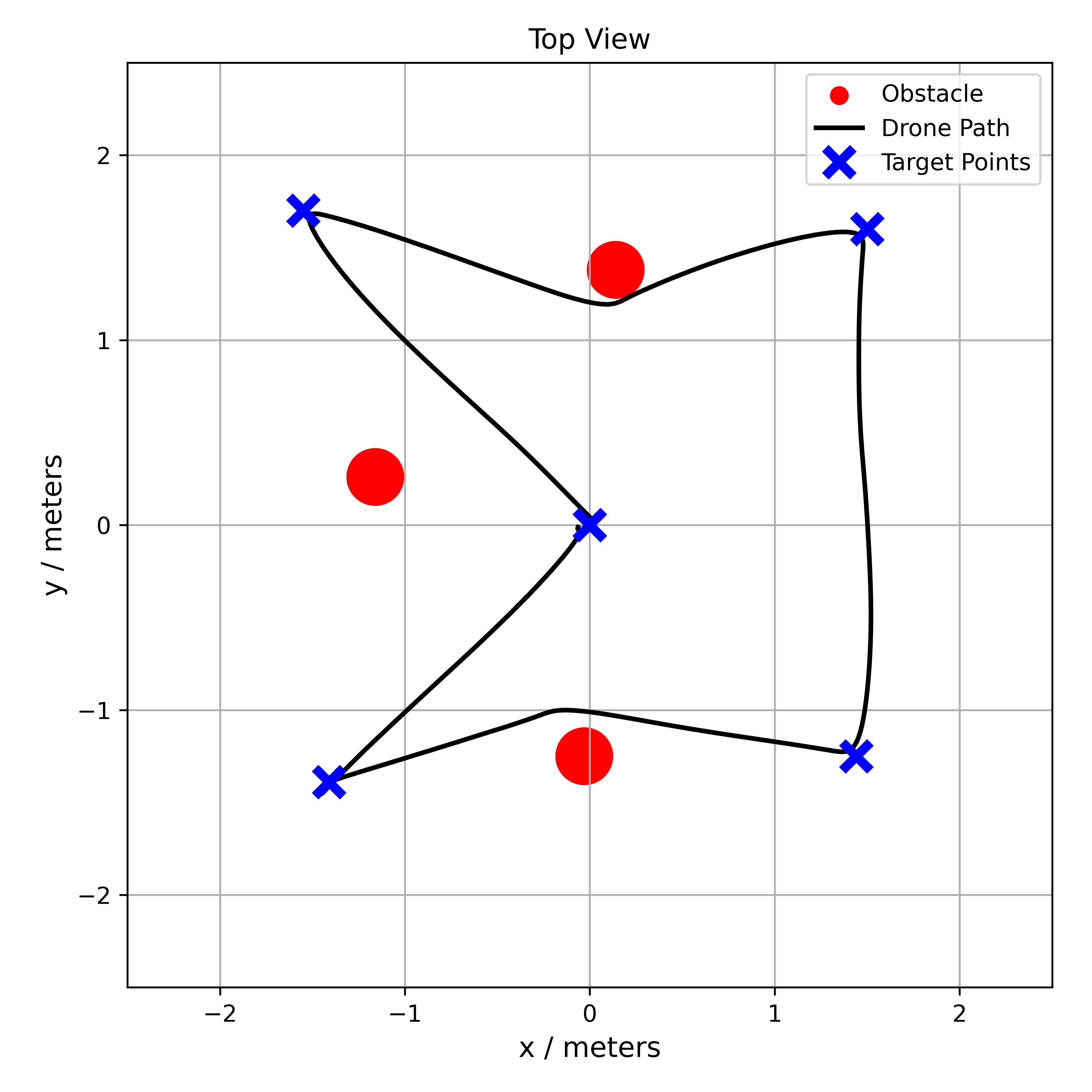}
  \caption{Scene 2 - Off-the-Shelf Autopilot.}
  \label{fig:sub2}
\end{subfigure}
\begin{subfigure}{.24\textwidth}
  \centering
  \includegraphics[width=1\linewidth]{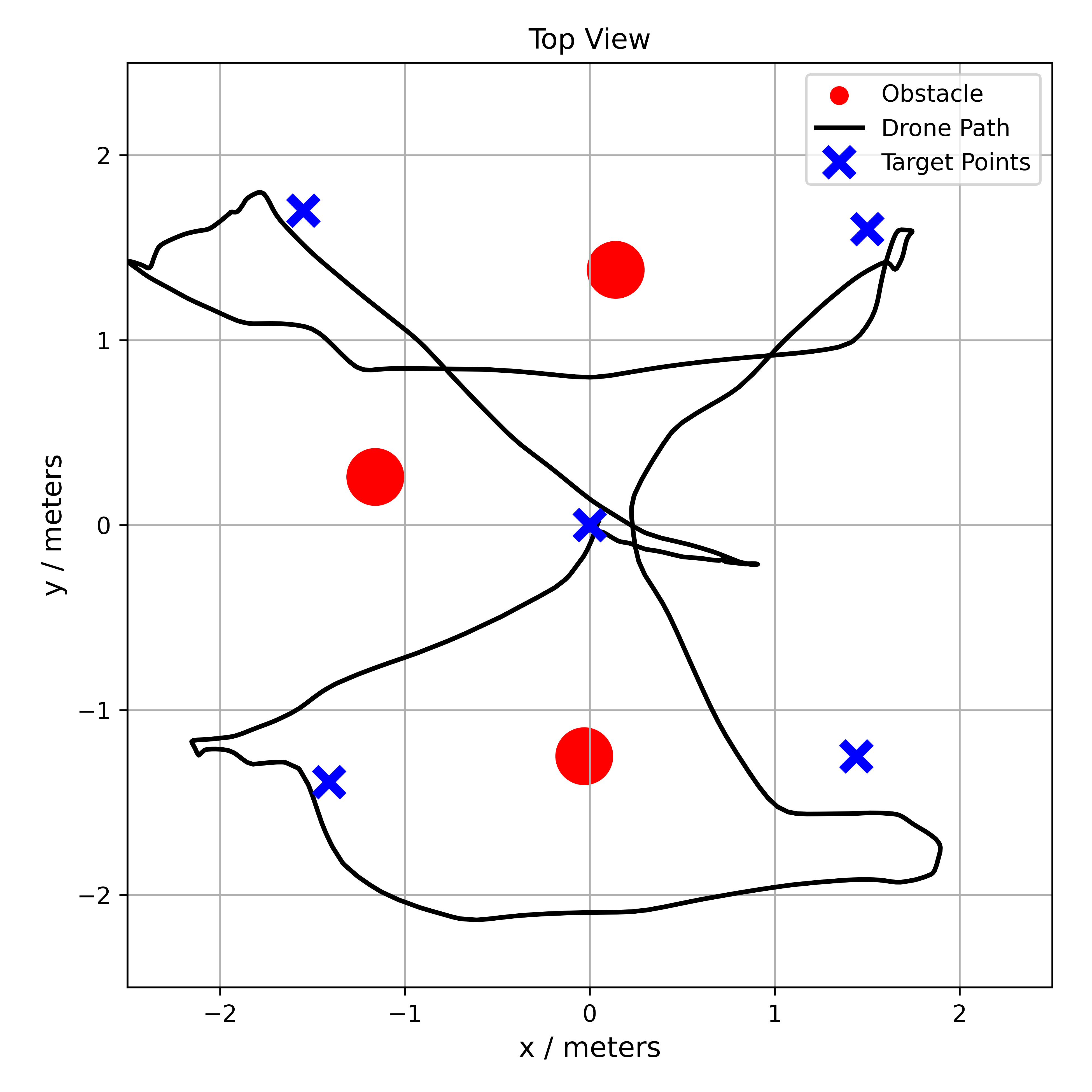}
  \caption{Scene 2 - Human Pilot.}
  \label{fig:sub2}
\end{subfigure}
\caption{Flight paths for the 3 experiments for Scene 2.}
\label{fig:scene2_results}
\end{figure}

\section{Conclusion}
In this work, we present the UAV-VLRR framework which is aimed at improving emergency response times in drone search and rescue operations. We demonstrate that our framework outperforms other traditional approaches in the field of drone search and rescue:

\begin{itemize}
    \item The text input provided a more natural way for the operator to design the search and rescue mission rather than observing the image and manually entering in waypoints.
    \item The point-to-point NMPC provided rapid response for quick mission completion.
    \item The amalgamation of these 2 aspects resulted in much shorter times for completion of missions.
    \item Our framework was tested on 2 different scenarios and was faster on an average by \textbf{33.75\%} when compared with off-the-shelf autopilot and \textbf{54.6\%} when compared with a human pilot.
\end{itemize}

These enhanced response times can be crucial in real-life scenarios where a matter of few seconds can prove to be very important.


\section{Future Work}
Future work on the UAV-VLRR system will focus on incorporating adaptive learning techniques to improve performance over time. In addition, exploring real-time coordination between multiple UAVs could enhance coverage and efficiency in large-scale SAR operations. 

Another important direction is the integration of dynamic environmental factors, such as moving obstacles, to improve the system’s relevance and robustness in real-world conditions. This will enable the UAVs to better navigate complex and unpredictable scenarios, which are common in search and rescue missions.

\section*{Acknowledgements} 
Research reported in this publication was financially supported by the RSF grant No. 24-41-02039.

\balance

\bibliographystyle{IEEEtran}
\bibliography{ref}

\end{document}